\documentclass{article}

\usepackage{arxiv}

\usepackage[utf8]{inputenc} 
\usepackage[T1]{fontenc}    
\usepackage{hyperref}       
\usepackage{url}            
\usepackage{booktabs}       
\usepackage{amsfonts}       
\usepackage{nicefrac}       
\usepackage{microtype}      
\usepackage{lipsum}
\usepackage{graphicx}
\usepackage{subfig}

\title{Haploid-Diploid Evolution: Nature's Memetic Algorithm}

\author{
    Michail-Antisthenis Tsompanas  \\
    Unconventional Computing Laboratory,\\
    University of the West of England,\\ 
    Bristol BS16 1QY, UK \\
    \texttt{antisthenis.tsompanas@uwe.ac.uk}
\And
       Larry Bull \\
     Department of Computer Science and \\Creative Technologies, \\ University of the West of England, \\ Bristol BS16 1QY, UK
\And
       Andrew Adamatzky \\
       Unconventional Computing Laboratory,\\
    University of the West of England,\\ 
    Bristol BS16 1QY, UK \\
\And
       Igor Balaz\\
       Laboratory for Meteorology, Physics and Biophysics,\\ Faculty of Agriculture, \\ Trg Dositeja Obradovica 8, \\University of Novi Sad, 21000, Novi Sad, Serbia
}

\begin{document}
\maketitle

\begin{abstract}
This paper uses a recent explanation for the fundamental haploid-diploid lifecycle of eukaryotic organisms to present a new memetic algorithm that differs from all previous known work using diploid representations. A form of the Baldwin effect has been identified as inherent to the evolutionary mechanisms of eukaryotes and a simplified version is presented here which maintains such behaviour. Using a well-known abstract tuneable model, it is shown that varying fitness landscape ruggedness varies the benefit of haploid-diploid algorithms. Moreover, the methodology is applied to optimise the targeted delivery of a therapeutic compound utilizing nano-particles to cancerous tumour cells with the multicellular simulator PhysiCell.
\end{abstract}


\section{Introduction}
\label{S:1}

The vast majority of work within evolutionary and memetic computation has used an underlying haploid representation scheme; individuals are each one solution to the given problem~\cite{eiben2003introduction}. Typically, bacteria contain one set of genes, whereas the more complex eukaryotic organisms --- such as plants and animals --- are predominantly diploid and contain two sets of genes~\cite{trun2009fundamental}. A small body of work exists using a diploid representation scheme within evolutionary computation; individuals each carry two solutions to the given problem~\cite{branke1999memory,branke2003designing,uyar2005new,fogel2006evolutionary}. In all but one known example, a dominance scheme is utilized to reduce the diploid down to a traditional haploid solution for evaluation. That is, as individuals carry two sets of genes, a heuristic is included to choose which of the genes to use in the evaluation process (see \cite{bhasin2016applicability} for a review). The only known exception is work by \cite{hillis1990co} on sorting networks wherein non-identical solution values at a given gene position are both used to form the solution for evaluation, thereby enabling solution lengths to vary up to the combined length of the two constituent genomes. 

\begin{figure*}[tbp]
\centering
\includegraphics[width=0.7\linewidth]{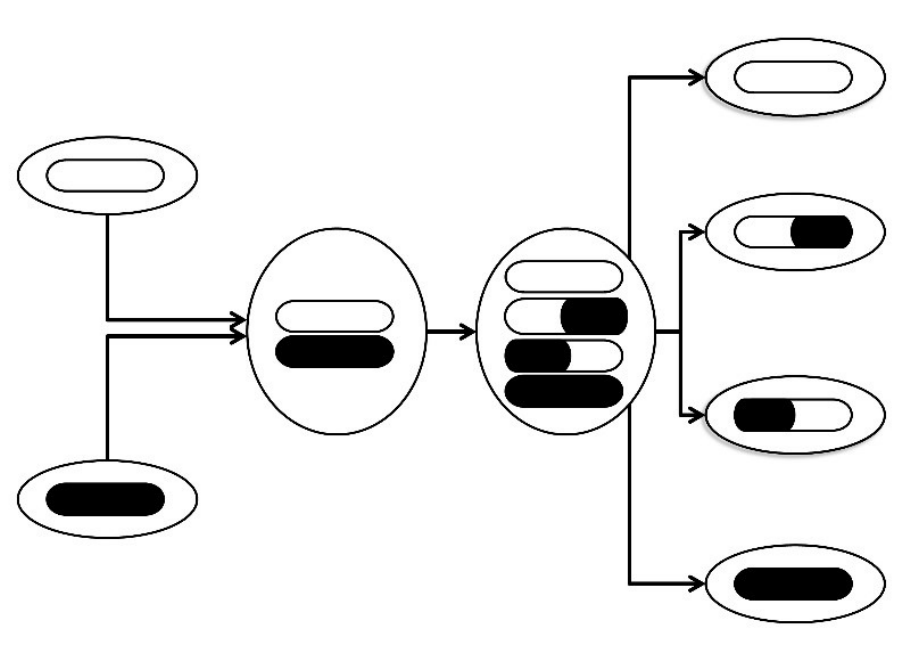}
\caption{Two-step meiosis with recombination under haploid-diploid reproduction as seen in most eukaryotic organisms (after \cite{smith1997major}).}
\label{fig:1}
\end{figure*}

In nature, eukaryotes exploit a haploid-diploid cycle, where one set of haploid cells from each of the diploid parents fuse to form a diploid cell/organism offspring. Specifically, each of the two genomes in an organism is replicated, with one copy of each genome being crossed over. In this way, copies of the original pair of genomes may be passed on, mutations aside, as might two versions containing a mixture of genes from each genome (Fig. \ref{fig:1}). Previous explanations for the emergence of the alternation between the haploid and diploid states in eukaryotes are typically based upon its being driven by changes in the environment (after \cite{margulis1986origins}). 

Recently, an explanation for the haploid-diploid cycle has been presented \cite{bull2017evolution}, which also explained other aspects of their sexual reproduction, including the use of recombination, based upon the Baldwin effect \cite{baldwin1896new}. The aim of this paper is to highlight the new understanding of the biological phenomenon and to present an example of a new general form of memetic computation, which similarly exploits a haploid-diploid cycle and, hence, a form of the Baldwin effect. Significantly, since the Baldwin effect is inherent to the search process, it does not preclude the inclusion of other local search mechanisms. 

The rest of the paper is arranged as follows: the next section presents the new understanding of how eukaryotic organisms evolve, a new simplified haploid-diploid algorithm is then presented, which maintains the basic mechanisms of the natural case. Finally, the new approach is applied to a high-throughput multicellular simulator to find potentially new therapeutic designs that maximise cancer tumour regression.

\section{Eukaryotic Evolution and the Baldwin effect}
\label{S:2}

The existence of phenotypic plasticity potentially enables an organism to display a different (better) fitness than its genome directly represents. Importantly, such ``learning'' can affect (improve) the evolutionary process by altering the shape of the underlying fitness landscape, such as smoothing over fitness valleys. For example, if a very poor combination of genes is able to consistently learn a fitter phenotype, their frequency will increase under selection more than expected without learning; the effective shape of the fitness landscape will be changed for the poor gene combination. As has been highlighted \cite{bull2017evolution}, becoming diploid can potentially alter gene expression in comparison to being haploid and, hence, affect the expected phenotype of each haploid alone, since both genomes are active in the cell --- through changes in gene product concentrations, partial or co-dominance, etc. That is, the fitness of the diploid cell/organism is a combination of the fitness contributions of the composite haploid genomes. If the cell/organism subsequently remains diploid and reproduces asexually, there is no scope for a rudimentary Baldwin effect. However, if there is a reversion to a haploid state for reproduction, there is the potential for a significant mismatch between the utility of the haploids passed on compared to that of the diploid selected; individual haploid gametes do not contain all of the genetic material through which their fitness was determined. That is, the effects of haploid genome combination into a diploid can be seen as a form of phenotypic plasticity for the individual haploid genomes before they revert to a solitary state during reproduction: joining together as a very simple learning. This suggests an increased benefit from the haploid-diploid cycle as landscape ruggedness increases as it is known that the most beneficial amount of learning under the Baldwin effect increases with the ruggedness of the fitness landscape \cite{bull1999baldwin,bull2017evolution}. 

Significantly, since the phenotype and, hence, fitness of a eukaryotic individual is a composite of its two haploid genomes, evolution can be seen to be evaluating a generalization over the typical fitnesses of solutions found between the two points represented by its constituent genomes: a single fitness value is assigned to two points in the haploid fitness landscape. Note this is in direct contrast to typical cases of bacteria --- and most evolutionary algorithms --- where individuals can be seen to represent a single point in the (haploid) fitness landscape only. Numerous explanations exist for the benefits of recombination in both natural (e.g., \cite{bernstein2010evolutionary}) and artificial systems (e.g., \cite{spears2013evolutionary}). The latter focusing solely upon haploid genomes and neither considering the potential Baldwin effect under the haploid-diploid cycle. The role of recombination becomes clear under the new view: recombination moves the current end points in the underlying haploid fitness space which define the generalization either closer together or further apart. That is, recombination adjusts the size of an area assigned a single fitness value, potentially enabling higher fitness regions to be more accurately identified over time. Moreover, recombination can also be seen to facilitate genetic assimilation within the simple form of the Baldwin effect. The pairing of haploid genomes is seen as a learning step with the fitness of a given haploid affected by the genes of its partner. If the pairing is beneficial and the diploid cell/organism is chosen under selection to reproduce, the recombination process brings an assortment of those partnered genes together into new haploid genomes (Fig. \ref{fig:1}). In this way the fitter allele values from the pair of partnered haploids may come to exist within individual haploids more quickly than the under mutation alone. Hence, in the emergence of more complex organisms, natural evolution appears to have discovered a more sophisticated approach to navigating their fitness landscapes. 

The Baldwin effect has long been used within evolutionary computation \cite{hinton1987learning}. This paper aims to show how the benefits of a haploid-diploid cycle can be exploited as a form of memetic computation. However, rather than just adopt nature’s diploid representation scheme under which an individual requires each haploid genome to be evaluated and their fitnesses combined in some way, eg, via their numerical average (as in \cite{bull2017evolution}),  a simpler scheme is proposed as explained in the following. This is first explored using the NK model of fitness landscapes.

\section{The NK Model}
\label{S:3}

The NK model \cite{kauffman1987towards} was introduced to allow the systematic study of various aspects of fitness landscapes (see \cite{kauffman1993origins} for an overview). In the standard model, the features of the fitness landscapes are specified by two parameters: $N$, the length of the genome; and $K$, the number of genes that has an effect on the fitness contribution of each (binary) gene. Thus, increasing $K$ with respect to $N$ increases the epistatic linkage, increasing the ruggedness of the fitness landscape. The increase in epistasis increases the number of optima, increases the steepness of their sides, and decreases their correlation. 

\begin{figure*}[tbp]
\centering
\includegraphics[width=0.7\linewidth]{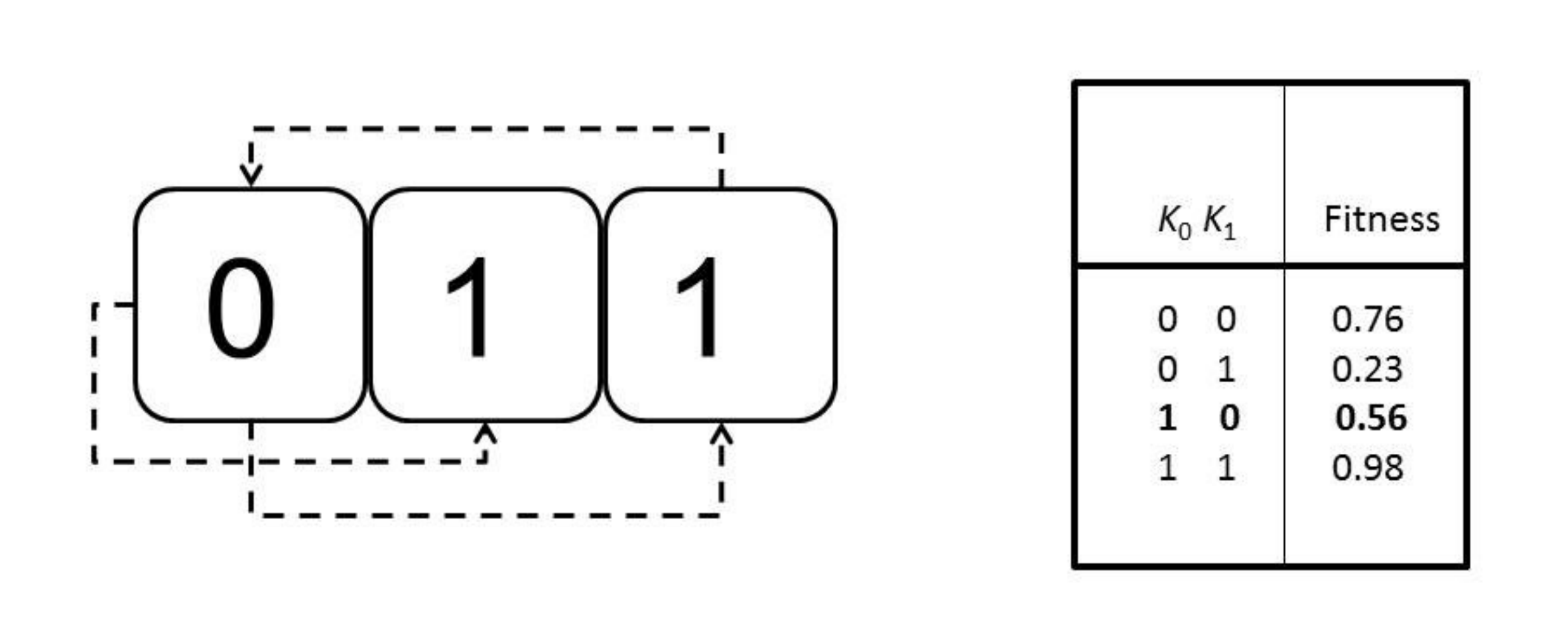}
\caption{An example NK model ($N$=3, $K$=1) showing how the fitness contribution of each gene depends on $K$ random genes (left). Therefore there are $2^{(K+1)}$ possible allele combinations per gene, each of which is assigned a random fitness. Each gene of the genome has such a table created for it (right, centre gene shown). Total fitness is the normalized sum of these values.}
\label{fig:2}
\end{figure*}

The model assumes all intragenome interactions are so complex that it is only appropriate to assign random values to their effects on fitness. Therefore for each of the possible $K$ interactions a table of $2^{(K+1)}$ fitnesses is created for each gene with all entries in the range 0.0 to 1.0, such that there is one fitness for each combination of traits (Fig. \ref{fig:2}). The fitness contribution of each gene is found from its table. These fitnesses are then summed and normalized by $N$ to give the selective fitness of the total genome. The results reported in the next section are the average of 10 runs (random start points) on each of 10 NK functions, i.e. 100 runs, for 20,000 generations. Here $0 \leq K \leq 15$, for $N=50$ and $N=100$.

\section{A Simple Haploid-Diploid Algorithm}
\label{S:4}

\begin{figure}[!tp]
\centering
    \subfloat[]{
      \includegraphics[width=.7\textwidth]{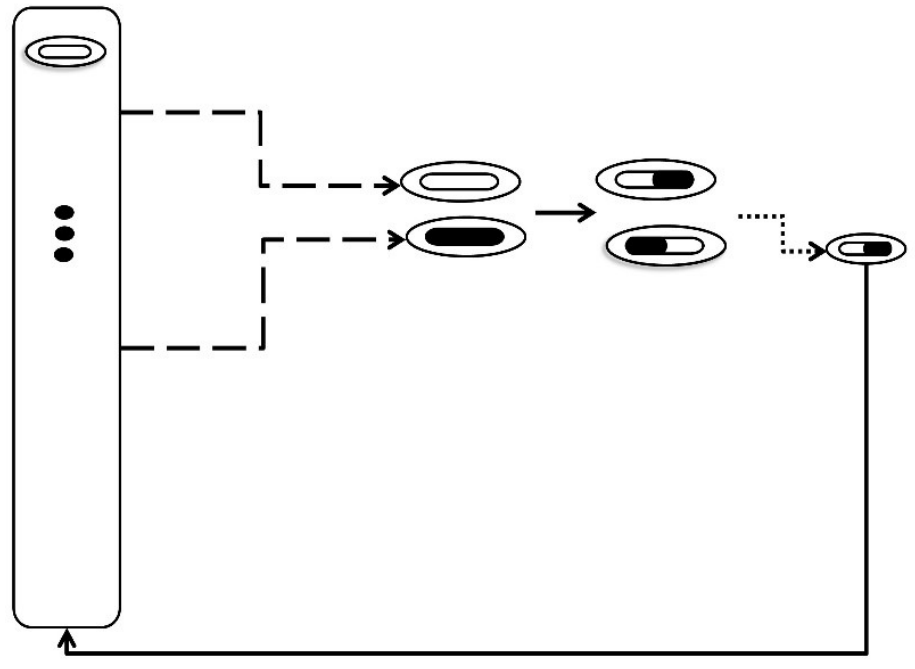}}
      
   \subfloat[]{
      \includegraphics[width=.7\textwidth]{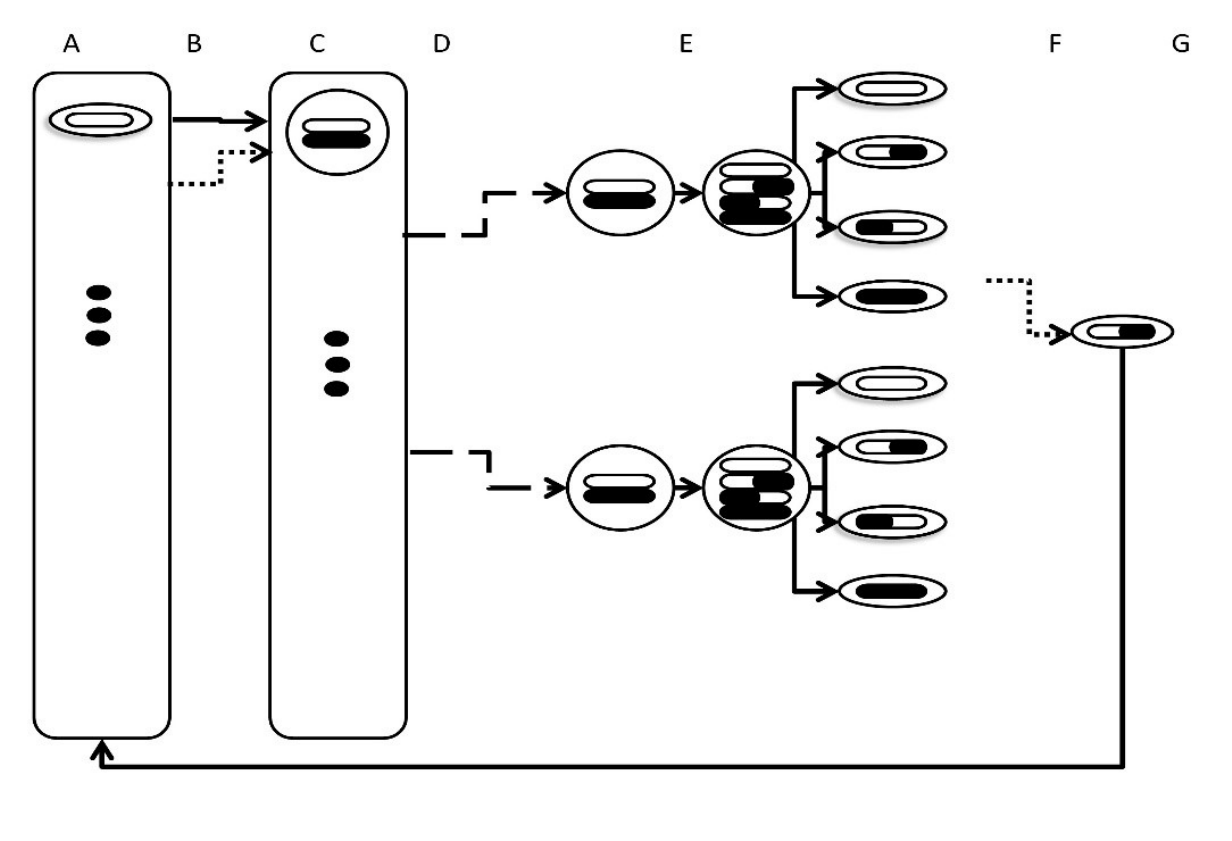}}
\caption{A schematic of the traditional evolutionary algorithm (a) and of the simple haploid-diploid algorithm (b).}
\label{fig:3}
\end{figure}

Figure \ref{fig:3}(a) shows a schematic of a traditional evolutionary algorithm (EA) which exploits one-point recombination, single-point mutation, and creates one offspring per cycle (steady state) which replaces the worst individual in the population here. Figure \ref{fig:3}(b) shows how the learning mechanism described above is implemented on top of that process. As can be seen: a traditional population of evaluated haploid individuals is maintained (A); a temporary population of diploid solutions is created from them by copying each haploid individual and then another haploid is chosen at random (B), with the fitness of the two haploids averaged (C); binary tournament selection then uses those fitnesses to pick two diploid parents (D); the haploid-diploid reproduction cycle with two-step meiosis as shown in Fig. \ref{fig:1} is then used for the two chosen parents (E); one of the resulting haploids is chosen at random, mutated, and evaluated (F); the offspring haploid is inserted into the original population (G).

\begin{figure}[htp]
\centering
    \subfloat[]{
      \includegraphics[width=.5\textwidth]{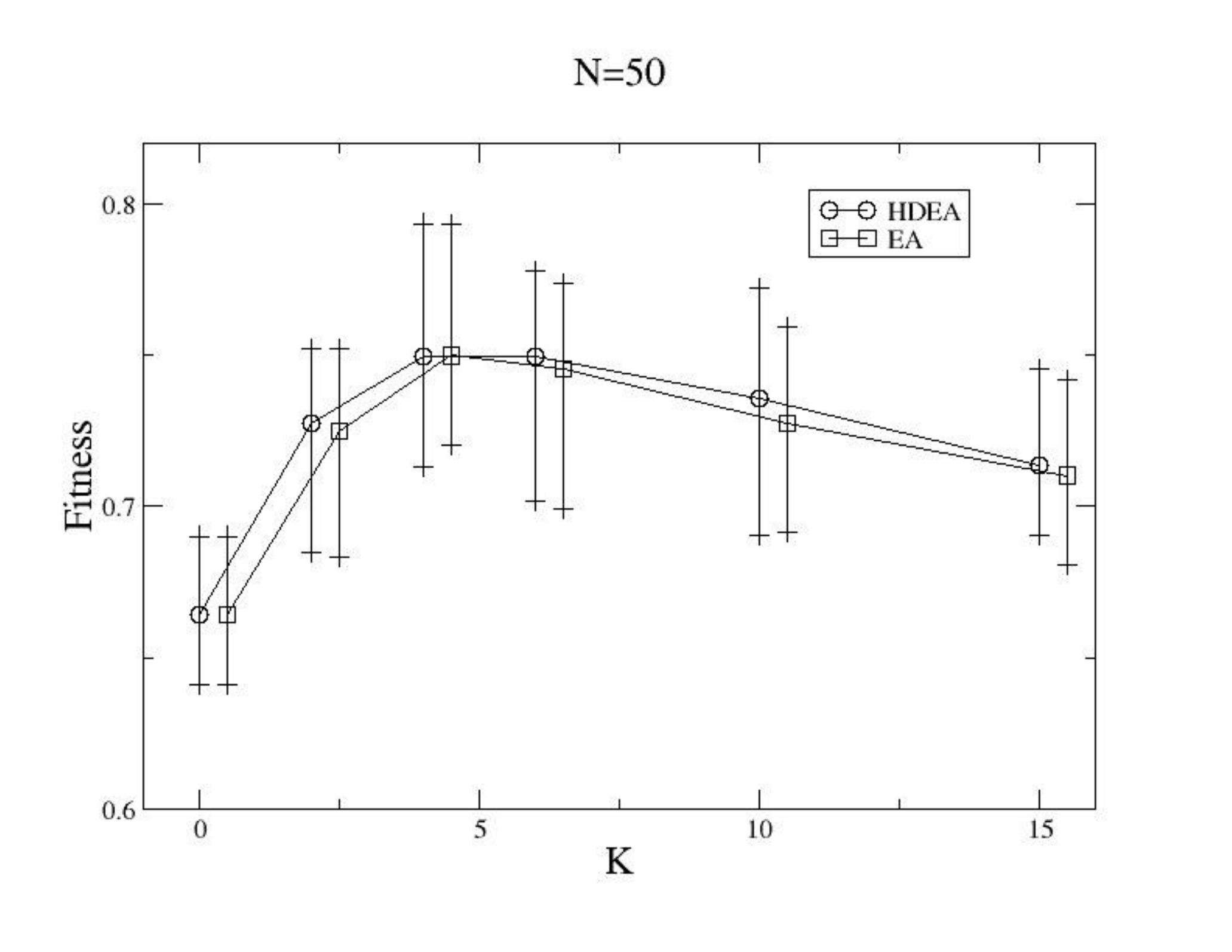}}
~
   \subfloat[]{
      \includegraphics[width=.5\textwidth]{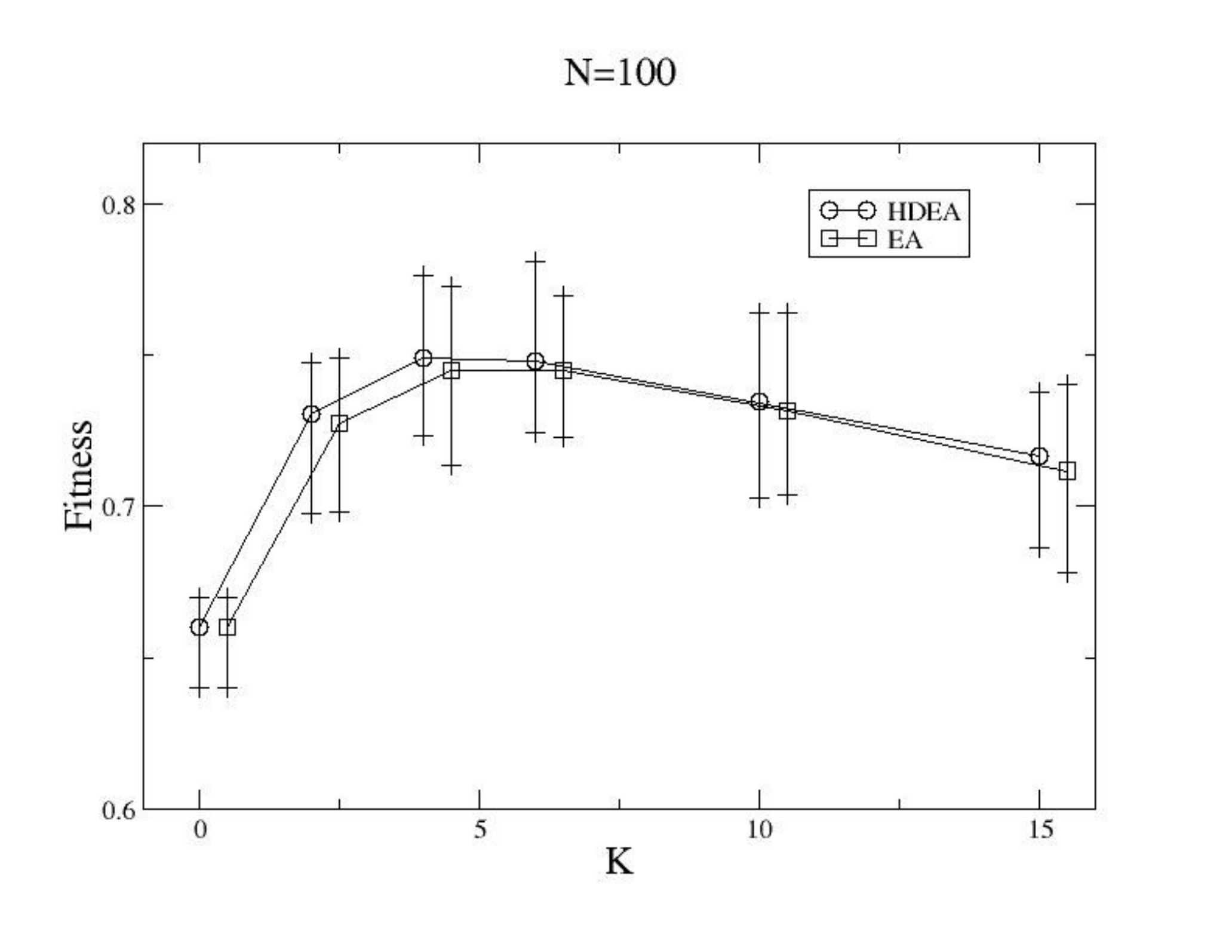}}
\caption{Showing examples of the fitness reached after 20,000 generations on landscapes of various size ($N$) and ruggedness ($K$). Error bars show min and max values.}
\label{fig:4}
\end{figure}

\begin{figure}[htp]
\centering
    \subfloat[]{
      \includegraphics[width=.5\textwidth]{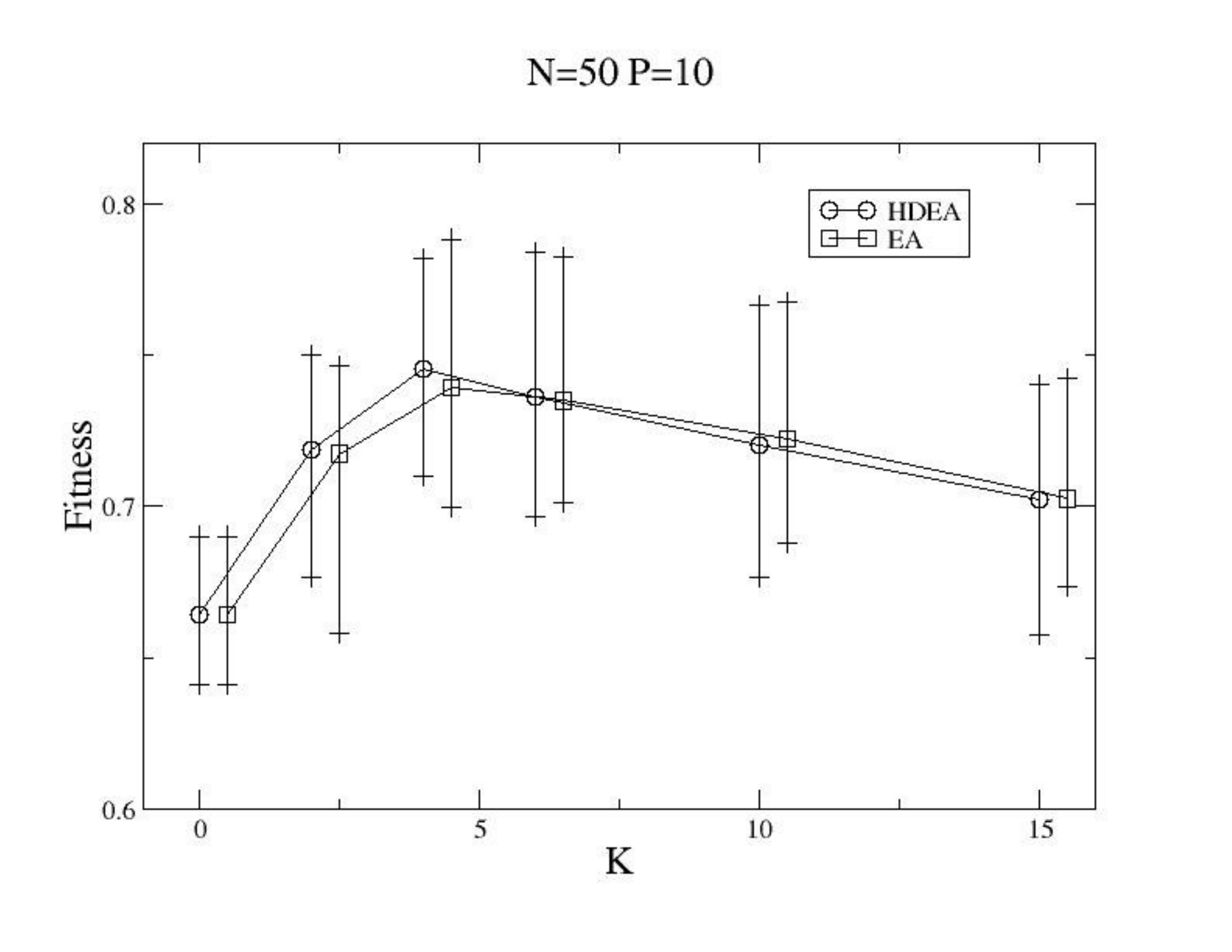}}
      ~
   \subfloat[]{
      \includegraphics[width=.5\textwidth]{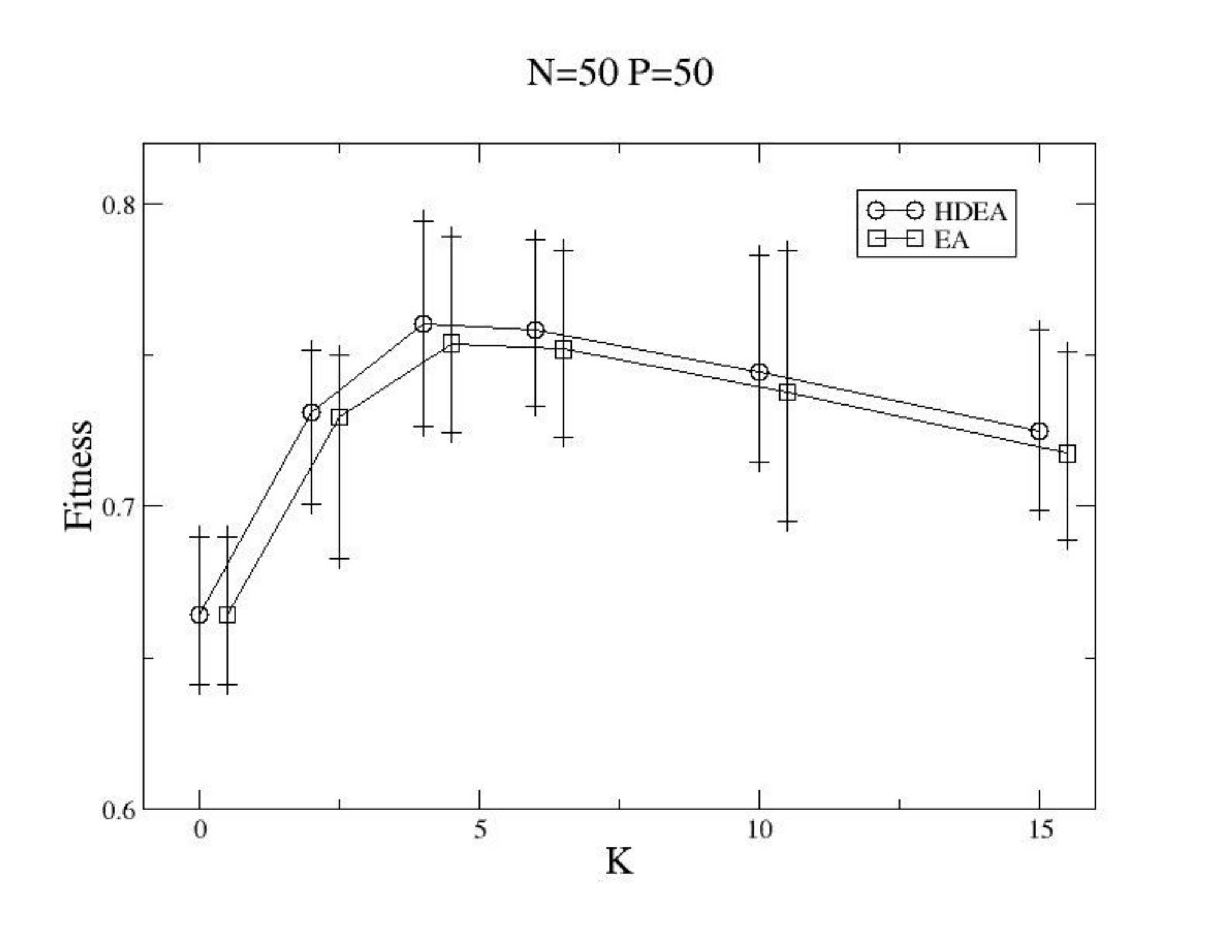}}
\caption{Showing examples of the fitness reached after 20,000 generations with differing population sizes ($P$).}
\label{fig:5}
\end{figure}

Figure \ref{fig:4} shows example results from running both the standard EA and the haploid-diploid EA (HDEA) on various NK fitness landscapes. Here population size $P=30$. As can be seen, when $K>4$, the HDEA performs best for $N=50$ and $K>2$ for $N=100$ (T-test, $p<0.05$). Thus, as anticipated, the simple Baldwin effect process proves beneficial with increased fitness landscape ruggedness. Figure \ref{fig:5} shows examples of how this is also true for different $P$, although the benefit is lost for higher $K$ when $P=10$. Related to this, since the HDEA makes a temporary population of diploids containing extra copies of randomly chosen haploid solutions, it might be argued that a larger population is available to selection than in the standard EA. Moreover, as the underlying traditional haploid population converges upon higher fitness solutions, the random sampling could be increasing their number and, thereby, altering the comparative selection pressure over time. However, results from simply creating a temporary haploid population of size $2P$, in the same way as the diploid population, does not alter performance significantly (not shown here, e.g., see \cite{karafotias2014parameter} for discussions of dynamic population sizing in general).

The findings with this abstract model are now explored in the context of simulating nano-particle therapy delivery for cancer tumour regression within PhysiCell v.1.5.1 (see \cite{metzcar2019review} for an overview of computational modelling in cancer biology).

\section{PhysiCell: A Physics-based Multicellular Simulator}
\label{S:5}

Among an increasing amount of computational models \cite{metzcar2019review} studying different aspects of cancer physiology, PhysiCell \cite{ghaffarizadeh2018physicell} is one of the leading ones. The open source simulator is based on a biotransport solver (BioFVM \cite{ghaffarizadeh2015biofvm}) and simulates a multicellular environment. While PhysiCell simulates cell cycling, death states, volume changes, mechanics, orientation and motility, it relies on BioFVM to simulate substrate secretion, diffusion, uptake, and decay. A significant advantage of PhysiCell is its open-source code that enables addition of new environmental substrates, cell types, and systems of cells, resulting in a general-purpose tool for investigating systems with multiple kinds of cells. This includes the ability to design cell-cell interaction rules to create a multicellular cargo delivery system that actively delivers a cancer therapeutic compound beyond regular drug transport limits to hypoxic cancer regions. We are currently exploring the use of evolutionary computing and other related techniques to optimise the design of such nano-particle (NP) delivery systems \cite{preen2019towards}. 

To evaluate the efficiency of the design of these NP delivery systems, the 2-D anti-cancer biorobots scenario of PhysiCell v.1.5.1
\cite{ghaffarizadeh2018physicell} was studied. This scenario utilizes three types of agents to simulate a high-throughput testing of a simple targeted drug delivery therapy. Namely, these types of agents are cancer cells, worker cells and cargo cells. Cancer cells consume oxygen and secrete a chemoattractant. The resulted gradient in oxygen concentration is employed to steer NPs, simulated as worker cells. These worker cells can be bonded with cargo cells, simulating the therapeutic compound. When a worker cell carries a cargo cell, it executes a random walk (migration) towards the gradient of the oxygen and, thus, towards accumulation of cancer cells. Whereas, when a worker cell does not carry a cargo cell it executes a random walk towards the area of the cargo cells. These random walks or migrations are controlled by input parameters of the simulator, in the range $[0,1]$, with 0 representing Brownian motion and 1 deterministic motion.

Finally, cargo cells simulating the therapeutic compound, can attract worker cells by exuding another simulated chemoattractant (which diffuses under BioFVM rules). As described before, worker cells can carry the cargo cells and deposit them in the affinity of cancer cells, resulting in apoptosis of these cells. The specific proximity is given by the parameter defined as cargo release $O_2$ threshold.

As per the initial example \cite{ghaffarizadeh2018physicell} and other relevant studies \cite{preen2019towards}, in the 2-D anti-cancer biorobots scenario an initial 200 $\mu m$ radius tumour is simulated to grow for 7 days. Then, 450 cargo cells  and 50 worker cells are added in a simulated vein close to the tumour. Note here that while in previous studies a random number of each type of cells with its mean as in the aforementioned was added, here we add exactly 450 and 50 cells for every simulation to alleviate one factor of stochasticity. The simulated drug delivery system is simulated for 3 more days and then the results are analyzed.

One paradigm of this simulation (whole 10 days) takes approximately 5 minutes of wall-clock time on an Intel\textregistered{}  Xeon\textregistered{} CPU E5-2650 at 2.20GHz with 64GB RAM using 8 of the 48 cores. To accelerate the computations and further alleviate the effect of the stochastic nature of the simulator on the results, a single tumour was used for testing every possible individual in the search space. For each test, one pre-grown tumour (for 7 days) was loaded to the simulator and the treatment was applied immediately. The test was finalized after 3 days from the introduction of the treatment, resulting in a minimization of wall-clock time to approximately 1,5 minutes. 
A static sampling approach is used, where the average of the outputs after 5 runs of the simulator with the same set of parameters was examined. The objective was determined as the remaining amount of cancer cells in the simulated area after the 3 days of treatment.

The search space was defined as a 6-dimensional space, with the 6 most prolific parameters for the behaviour of worker cells (or simulated NPs). Namely, the parameters under investigation were: the attached worker migration bias [0,1]; the unattached worker migration bias [0,1]; worker relative adhesion [0,10]; worker relative repulsion [0,10]; worker motility persistence time (minutes) [0,10]; and the cargo release $O_2$ threshold ($mmHg$) [0,20]. The rest of the parameters on the simulator are not altered from the initial distribution of the simulator \cite{ghaffarizadeh2018physicell} and depicted in Table \ref{tabl:1}.

\begin{table}
\centering
\caption{Unaltered parameters of PhysiCell simulator.}
\label{tabl:1}       
\begin{tabular}{rl}
\hline\noalign{\smallskip}
Parameter & Value  \\
\noalign{\smallskip}\hline\noalign{\smallskip}
Maximum attachment distance & 18 $\mu m$ \\
Minimum attachment distance & 14 $\mu m$ \\ 
Worker apoptosis rate      & 0 $min^{-1}$ \\ 
Worker migration speed     & 2 $\mu m / min$ \\ 
Worker $O_2$ relative uptake  & 0.1 $min^{-1}$ \\ 
Cargo $O_2$ relative uptake   & 0.1 $min^{-1}$ \\
Cargo apoptosis rate       & 4.065e-5 $min^{-1}$ \\ 
Maximum relative cell adhesion distance & 1.25 \\ 
Maximum elastic displacement & 50 $\mu m$  \\
Damage rate &  0.03333 $min^{-1}$\\
Repair rate &  0.004167 $min^{-1}$\\
Drug death rate &  0.004167 $min^{-1}$\\
Cargo relative adhesion    & 0  \\
Cargo relative repulsion   & 5  \\
Elastic coefficient &  0.05 $min^{-1}$  \\
Motility shutdown detection threshold &  0.001 \\
Attachment receptor threshold  & 0.1 \\

\noalign{\smallskip}\hline
\end{tabular}
\end{table}

\section{Results of HDEA optimization on PhysiCell}
\label{S:5a}

\begin{figure*}[tbp]
\centering
\includegraphics[width=0.6\linewidth]{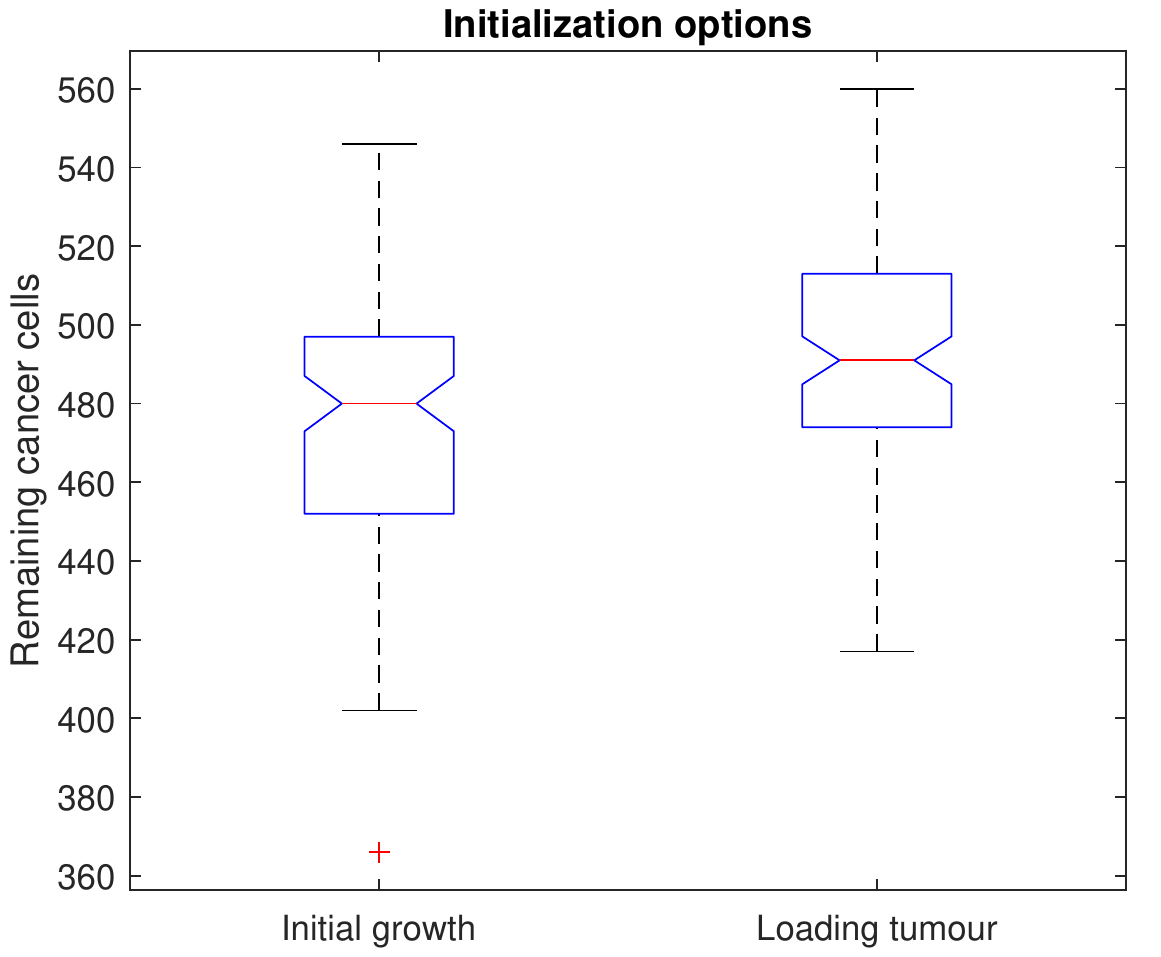}
\caption{Boxplot of 100 samples for each of the initialization options.}
\label{fig:boxInit}
\end{figure*}

Initially the option to load a tumour rather than simulate its growth for 7 days was investigated. In Fig. \ref{fig:boxInit} the boxplot of 100 simulations for each initialization option with the same input parameters is illustrated. When comparing the initial growth (7 days tumour growth and 3 days treatment simulation, mean=475.06, SD=32.9, median=480, kurtosis=3.3515) with the loading tumour alternative (loading a tumour and 3 days treatment simulation, mean=494.12, SD=29.11, median=491, kurtosis=2.7698), the latter produces more consistent results (based on smaller standard deviation and kurtosis). Additional to the aforementioned acceleration of computations (from 5 minutes to 1,5 minutes) the loading tumour was selected for the tests presented in the following.

To study the performance of HDEA, another control algorithm was utilized to optimize the behaviour/design of worker cells, namely a steady-state genetic EA. The population size was set to $P=50$, the selection and replacement tournament size to $T=3$, a uniform crossover probability to $X=80\%$ and a per allele mutation rate to $\mu=20\%$ with a uniform random step size of range $s=[-5,5]\%$. The HDEA was set up with the same parameters as the EA in order for the comparison to be meaningful. All comparison runs started by evaluating a randomly produced, same for each run, initial population ($P=50$) under PhysiCell simulator, and then using the corresponding EA to evolve the design of worker cells, with a computational budget of 100 individual evaluations (100 individuals $\times$ 5 samples = 500 PhysiCell simulations). In total, 30 comparison runs were executed.


      
   

\begin{figure*}[tbp]
\centering
\includegraphics[width=0.6\linewidth]{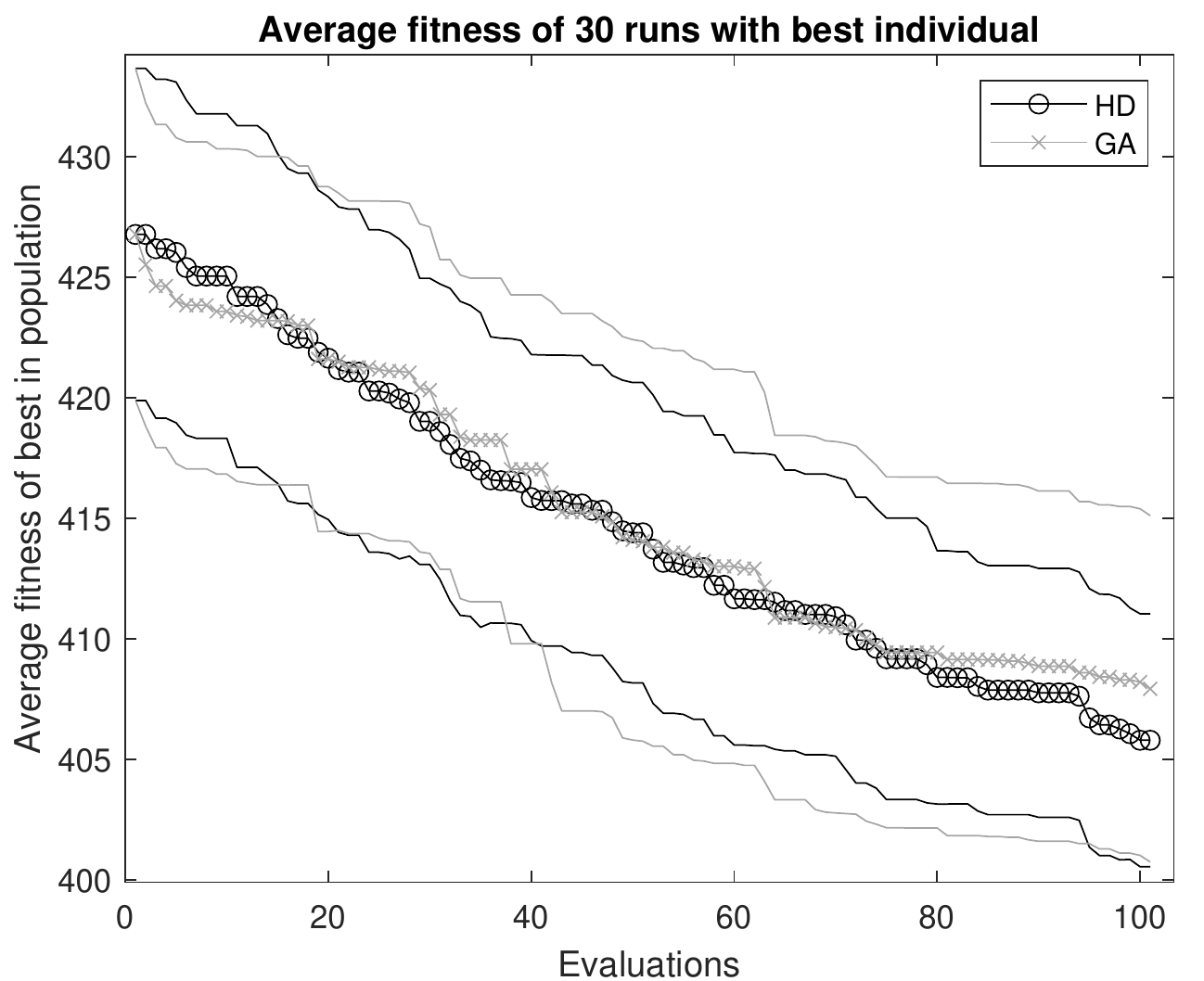}
\caption{Average and confidence levels (95\%) of the best individuals per evolution step for both algorithms for all 30 runs.}
\label{fig:totBest}
\end{figure*}

In Fig. \ref{fig:totBest} the evolution of the best individuals found by the two algorithms is illustrated. Specifically, the average and confidence level at 95\% for the best individuals in all 30 runs are considered. Throughout the evolution steps it is apparent that the HDEA algorithm is generally finding better solutions faster (it learns faster). Moreover, the final average of best solutions found by HDEA is better than the one by the genetic algorithm. The smaller range of the 95\% confidence levels of HDEA reveal a better consistency in the solutions found by this algorithm.

\begin{figure*}[tbp]
\centering
\includegraphics[width=0.6\linewidth]{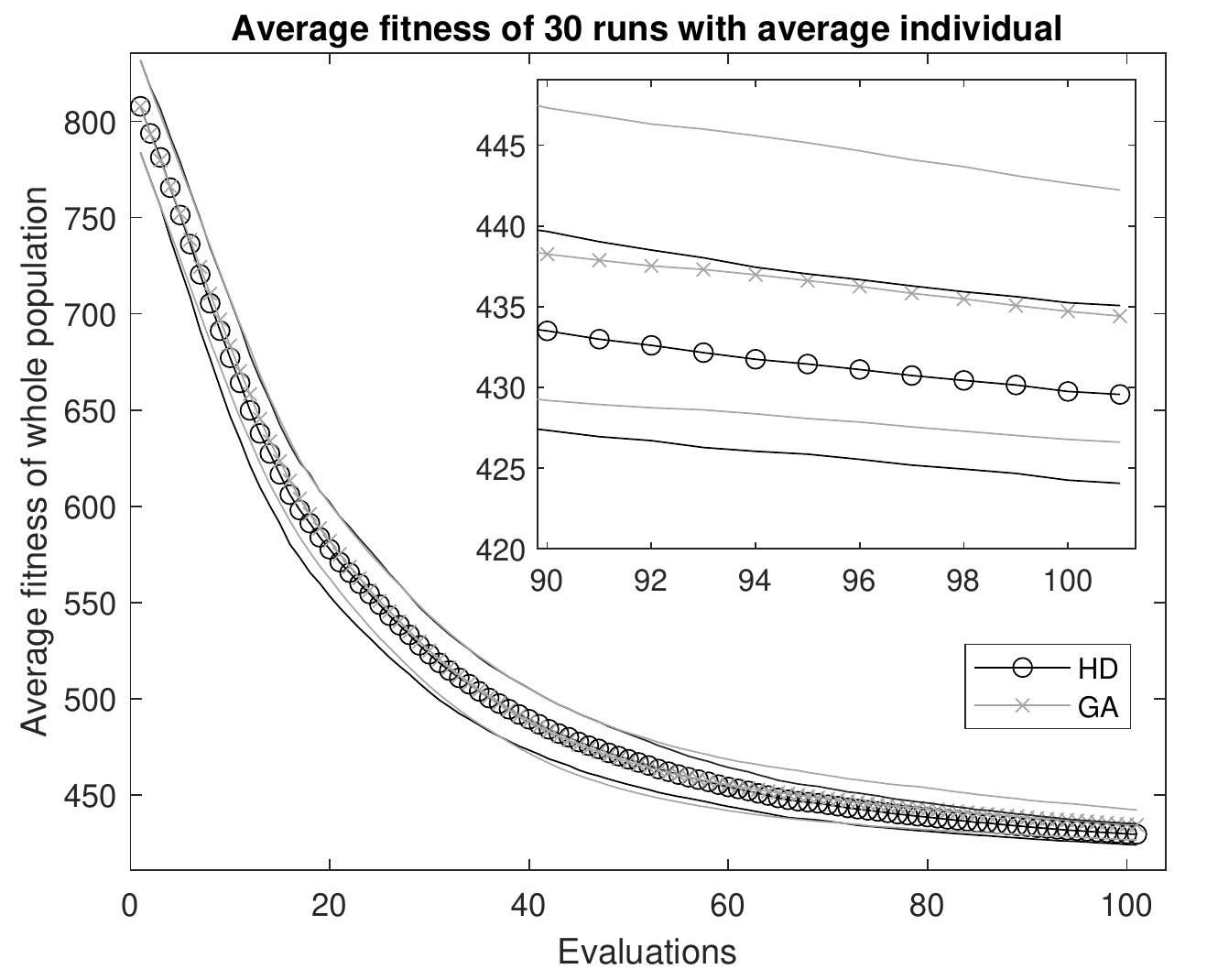}
\caption{Average and confidence levels (95\%) of all the individuals per evolution step for both algorithms for all 30 runs.}
\label{fig:totAve}
\end{figure*}

Figure \ref{fig:totAve} shows the relative performance of the average solutions over time for both approaches. As can be seen, the HDEA finds fitter solutions. Note the zoomed in region of evaluations 90 to 100 for a clearer comparison. Although, after 100 evaluations, the best solutions (Fig. \ref{fig:totBest}) are not statistically significantly better (Wilcoxon signed-rank test, $p=0.3763$), the average solutions (Fig. \ref{fig:totAve}) are (Wilcoxon signed-rank test, $p=0.0256$). It can be noted that the best solution found by the HDEA was significantly better (Wilcoxon signed-rank test, $p=0.0215$) for the first ten runs of the thirty shown here.

\begin{figure*}[tbp]
\centering
\includegraphics[width=0.6\linewidth]{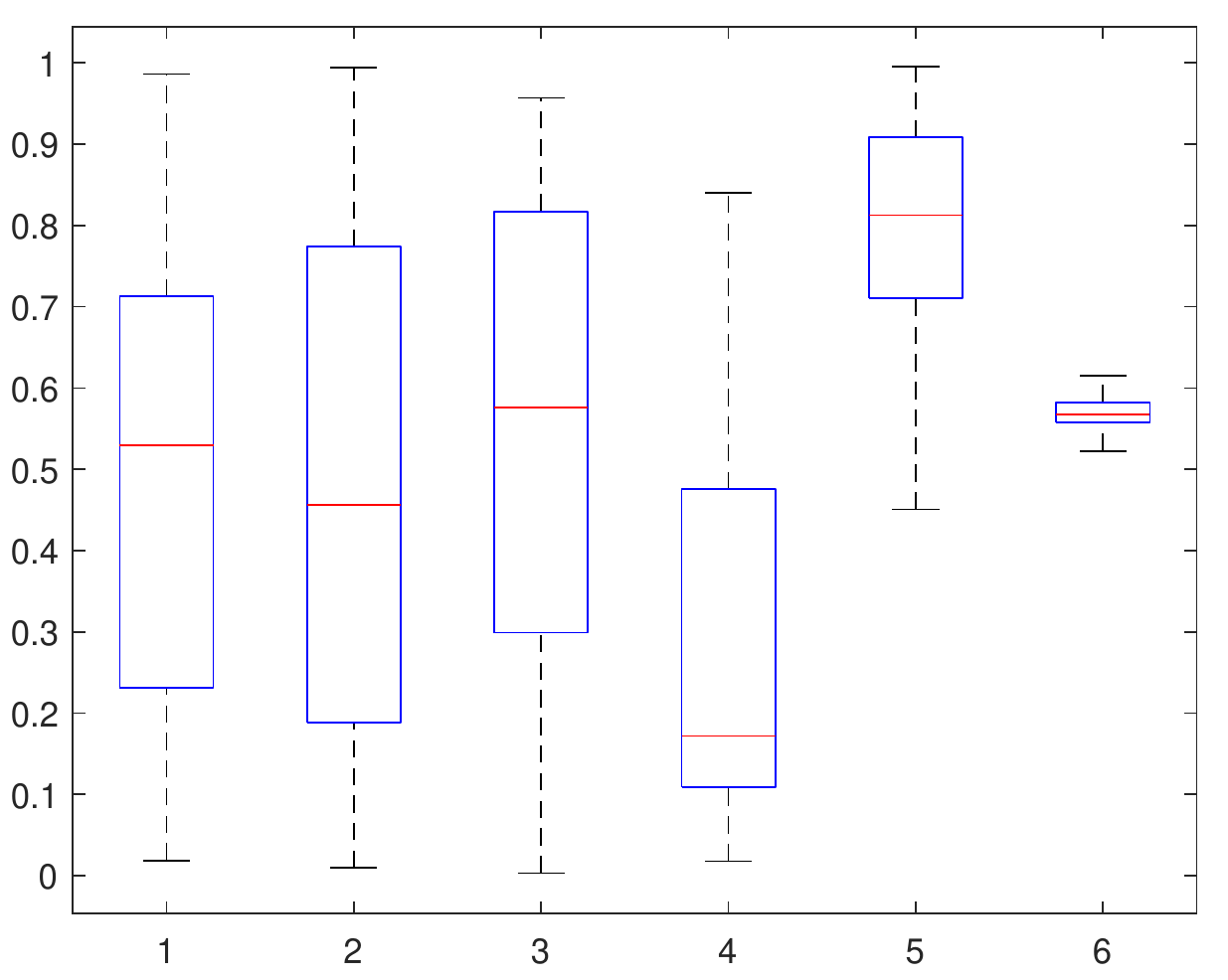}
\caption{Boxplot of parameters of best individuals found by GA (in normalized ranges). Parameters: 1) attached worker migration bias 2) the unattached worker migration bias 3) worker relative adhesion 4) worker relative repulsion 5) worker motility persistence time 6) and the cargo release $O_2$ threshold.}
\label{fig:GAbox}
\end{figure*}

\begin{figure*}[tbp]
\centering
\includegraphics[width=0.6\linewidth]{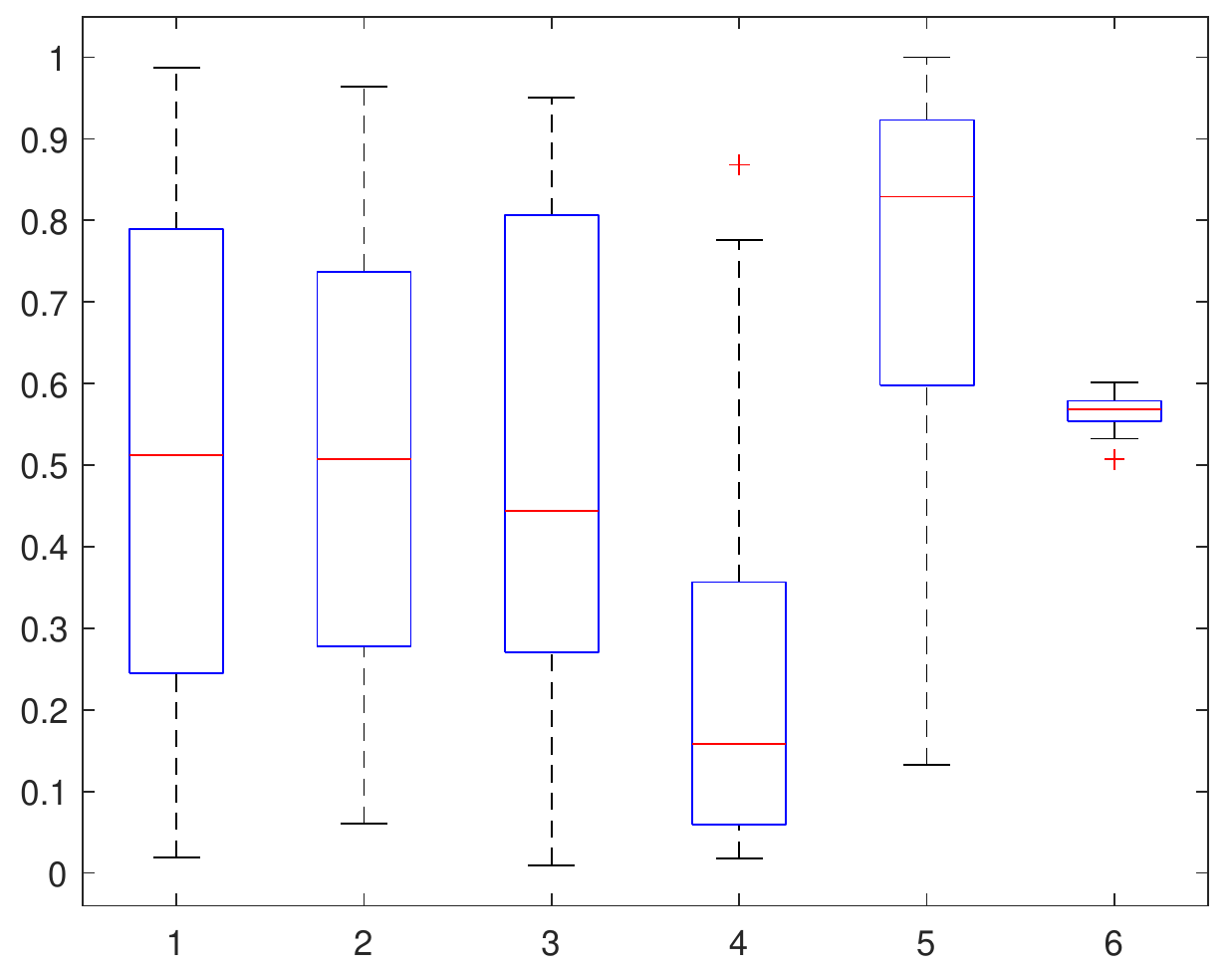}
\caption{Boxplot of parameters of best individuals found by HDEA (in normalized ranges). Parameters: 1) attached worker migration bias 2) the unattached worker migration bias 3) worker relative adhesion 4) worker relative repulsion 5) worker motility persistence time 6) and the cargo release $O_2$ threshold.}
\label{fig:HDbox}
\end{figure*}

\begin{figure*}[tbp]
\centering
\includegraphics[width=0.8\linewidth]{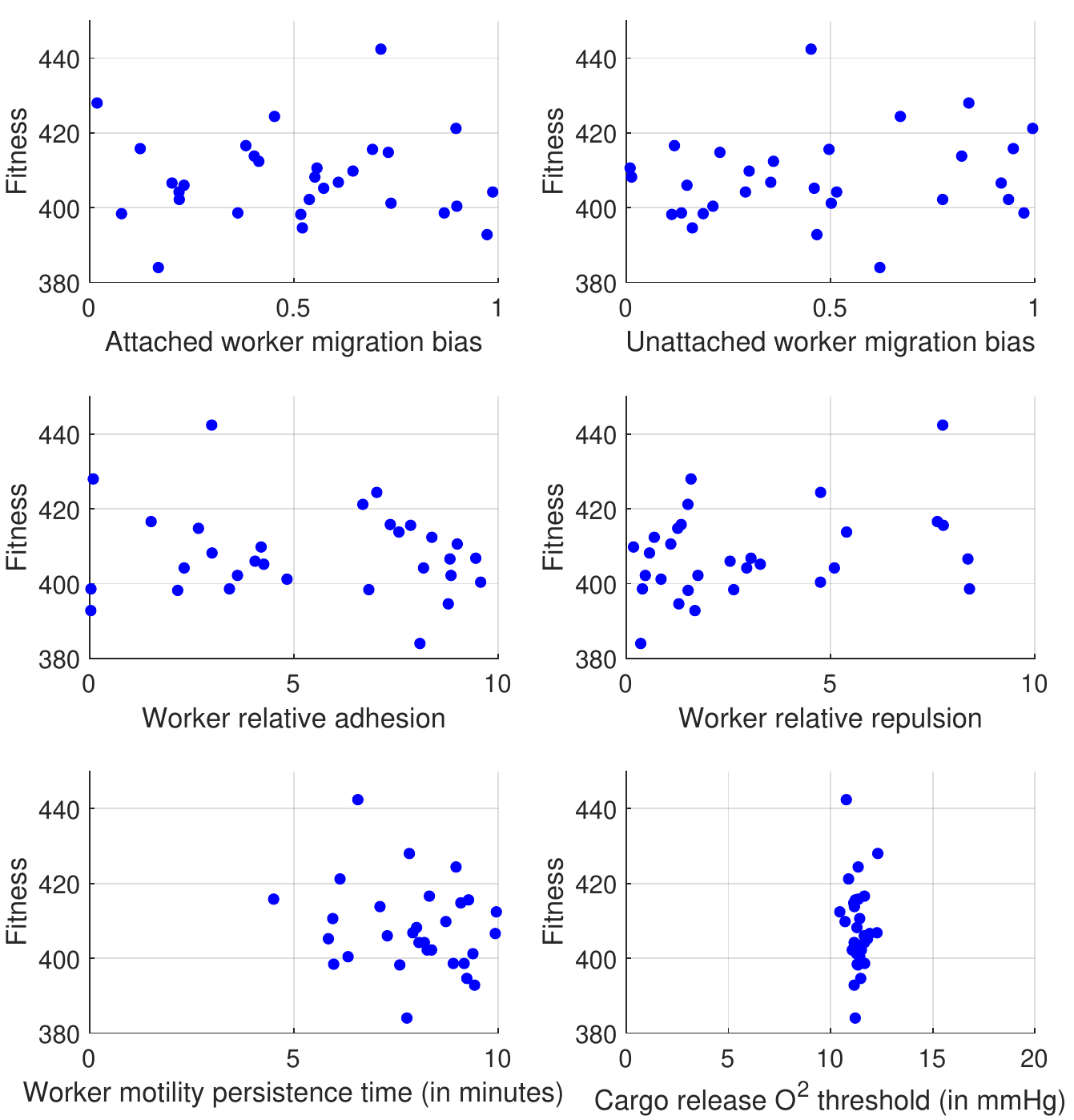}
\caption{Scatter plot of parameters of best individuals found by GA for all 30 runs.}
\label{fig:GAscat}
\end{figure*}

\begin{figure*}[tbp]
\centering
\includegraphics[width=0.8\linewidth]{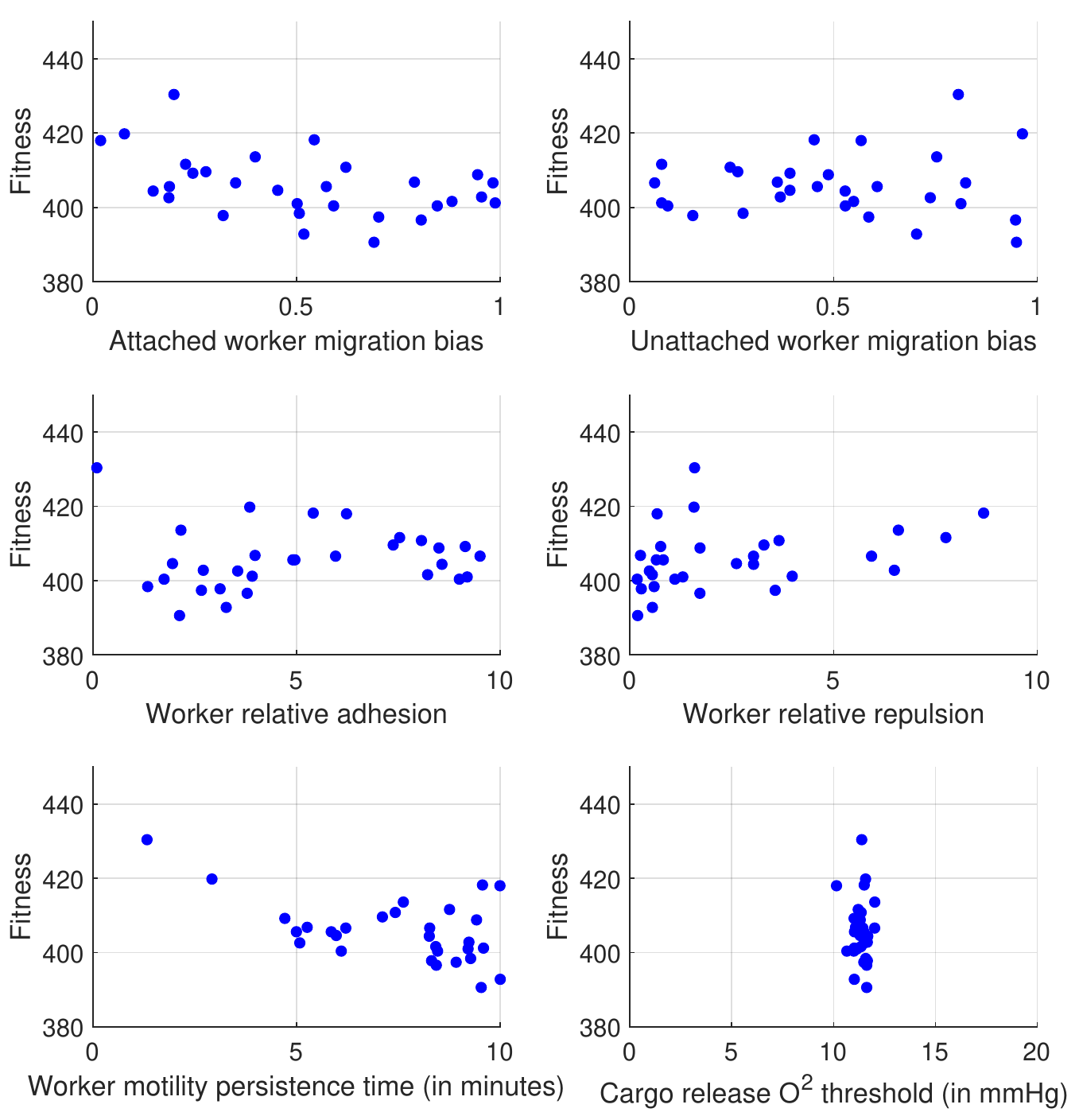}
\caption{Scatter plot of parameters of best individuals found by HDEA for all 30 runs.}
\label{fig:HDscat}
\end{figure*}

In Figs. \ref{fig:GAbox} and \ref{fig:HDbox} the boxplots of the parameters of the best individual discovered during the 30 runs by GA and HDEA, respectively, are presented. In Figs. \ref{fig:GAscat} and \ref{fig:HDscat} the scatter plots of the parameters of the best individual discovered during the 30 runs by GA and HDEA, respectively, are depicted. It is clear that the most prolific parameter value for optimizing the design of NPs is the cargo release $O_2$ threshold parameter. The majority of solutions are quite close to 11 $mmHg$, similar to findings from previous works \cite{ghaffarizadeh2018physicell,preen2019towards}. Although, for three of the parameters the results can not be conclusive (namely, attached and unatached worker migration bias and worker relative adhesion having almost uniform distribution like boxplots), the graphs for the other two parameters can convey the fact of the solutions being skewed towards smaller values for worker relative repulsion and higher values for worker motility persistence time. 

\section{Conclusion}
\label{S:6}

In the standard evolutionary computing approach each individual solution can be seen to represent a single point in the fitness landscape. Typically, the same is true of bacteria in natural evolution. It has recently been suggested that natural evolution is using a more sophisticated approach with eukaryotes, exploiting a generalization process, whereby each individual represents a region in the fitness landscape \cite{bull2017evolution}. Of course, landscape smoothing can be achieved by numerous mechanisms (after \cite{hinton1987learning}) but they all require extra fitness evaluations. The scheme presented in this paper is intended to exploit the Baldwin effect through what is essentially simple population manipulation rather than through altering the underlying representation and evaluations of the standard evolutionary computing approach.

It can also be noted that the shape of the fitness landscape varies based upon the haploid genomes, which exist within a given population at any time and how they are paired. This is significant since, as has been pointed out for coevolutionary fitness landscapes \cite{bull2001coevolutionary}, such movement potentially enables the temporary creation of neutral paths, where the benefits of (static) landscape neutrality are well-established (after \cite{kimura1983neutral}).

The proposed HDEA method was compared with a simple haploid EA in both an abstract model (NK model) and a complicated simulator (PhysiCell). In both cases HDEA seems to perform better than the traditional and well-established haploid method. Results from the the abstract model unveil that when the methodology is applied in problems with increased ruggedness in their fitness landscape, it performs better. After analyzing the results of the methodology on the cancer treatment simulator, it can be concluded that it reaches fitter solutions faster, despite the high stochasticity injected into the fitness landscape by the simulator.

\section{Acknowledgements}
This work was supported by the European Research Council under the European Union's Horizon 2020 research and innovation programme under grant agreement No. 800983.

\bibliographystyle{unsrt}

\end{document}